\documentclass{article} 
\usepackage{iclr2024_conference,times}

\usepackage{amsmath,amsfonts,bm}

\def\eqref#1{equation~\ref{#1}}

\def\1{\bm{1}}

\DeclareMathAlphabet{\mathsfit}{\encodingdefault}{\sfdefault}{m}{sl}
\SetMathAlphabet{\mathsfit}{bold}{\encodingdefault}{\sfdefault}{bx}{n}

\usepackage[colorlinks=true]{hyperref}
\usepackage{url}

\usepackage{booktabs}       
\usepackage{amsfonts}       
\usepackage{nicefrac}       
\usepackage{microtype}      
\usepackage{xcolor}         
\usepackage{wrapfig}
\usepackage{graphicx}    
\usepackage{multirow}
\usepackage{subcaption}

\title{MiniGPT-4:\\ Enhancing Vision-Language Understanding with Advanced Large Language Models}

\author{Deyao Zhu$^*$, Jun Chen\footnote{equal contribution},  Xiaoqian Shen, Xiang Li, Mohamed Elhoseiny \\
King Abdullah University of Science and Technology\\
\texttt{\{deyao.zhu,jun.chen,xiaoqian.shen,}\\
\texttt{xiang.li.1,mohamed.elhoseiny\}@kaust.edu.sa}
}

\iclrfinalcopy 
\begin{document}

\maketitle

\begin{abstract}
The recent GPT-4 has demonstrated extraordinary multi-modal abilities, such as directly generating websites from handwritten text and identifying humorous elements within images. These features are rarely observed in previous vision-language models. However, the technical details behind GPT-4 continue to remain undisclosed.
We believe that the enhanced multi-modal generation capabilities of GPT-4 stem from the utilization of sophisticated large language models (LLM). 
To examine this phenomenon, we present MiniGPT-4, which aligns a frozen visual encoder with a frozen advanced LLM, Vicuna, using one projection layer. 
Our work, for the first time, uncovers that properly aligning the visual features with an advanced large language model can possess numerous advanced multi-modal abilities demonstrated by GPT-4, 
such as detailed image description generation and website creation from hand-drawn drafts.
Furthermore, we also observe other emerging capabilities in MiniGPT-4, including writing stories and poems inspired by given images, teaching users how to cook based on food photos, and so on. 
In our experiment, we found that the model trained on short image caption pairs could produce unnatural language outputs (e.g., repetition and fragmentation). To address this problem, we curate a detailed image description dataset in the second stage to finetune the model, which consequently improves the model's generation reliability and overall usability. 
Our code, pre-trained model, and collected dataset are available at \url{https://minigpt-4.github.io/}.
\end{abstract}

\section{Introduction}
In recent years, large language models (LLMs) have experienced rapid advancements~\citep{instructGPT, chatgpt, gpt3, bloom, llama, chowdhery2022palm, hoffmann2022training}. With exceptional language understanding capabilities, these models can perform a variety of intricate linguistic tasks in a zero-shot manner. Notably, GPT-4, a large-scale multimodal model, has been recently introduced and demonstrated several impressive capabilities of vision-language understanding and generation~\citep{gpt4}. For example, GPT-4 can produce detailed and accurate image descriptions, explain unusual visual phenomena, and even construct websites based on handwritten text instructions.

Although GPT-4 has exhibited remarkable vision language capabilities, 
the methods behind its 
exceptional abilities are still a mystery \citep{gpt4}. 
We believe that these impressive skills may stem from the utilization of a more advanced large language model (LLM). 
LLMs have demonstrated various emergent abilities, as evidenced in GPT-3's few-shot prompting setup~\citep{gpt3} and the findings of Wei \textit{et al}. (2022)~\citep{wei2022emergent}. Such emergent properties are hard to find in smaller-scale models. It is conjectured that these emergent abilities are also applicable to multi-modal models, which could be the foundation of GPT-4's impressive visual description capabilities.

To substantiate our hypothesis, we present a novel vision-language model named MiniGPT-4. 
It utilizes an advanced large language model (LLM), Vicuna~\citep{vicuna2023}, which is built upon LLaMA~\citep{llama} and reported to achieve 90\% of ChatGPT's quality as per GPT-4's evaluation, as the language decoder. 
In terms of visual perception, we employ the same pretrained vision components of BLIP-2 \citep{blip2} that consists of a ViT-G/14 from EVA-CLIP~\citep{fang2022eva} and a Q-Former network. 
MiniGPT-4 adds a single projection layer to align the encoded visual features with the Vicuna language model and freezes all the other vision and language components. 
MiniGPT-4 is initially trained for 20k steps using a batch size of 256 on 4 A100 GPUs, leveraging a combined image captioning dataset that includes images from LAION~\citep{laion}, Conceptual Captions~\citep{changpinyo2021conceptual,sharma2018conceptual}, and SBU~\citep{ordonez2011im2text} to align visual features with the Vicuna language model.
Nevertheless, merely aligning visual features with the language model (LLM) is inadequate to ensure robust visual conversation capabilities, resembling that of a chatbot. 
The presence of underlying noise in raw image-text pairs can lead to subpar language outputs. 
Therefore, we collect another ~3,500 detailed image description pairs to further fine-tune the model with a designed conversational template in order to improve the naturalness of the generated language and its usability.

\begin{figure}[t]
\centering
 \includegraphics[width=0.75\linewidth]
    {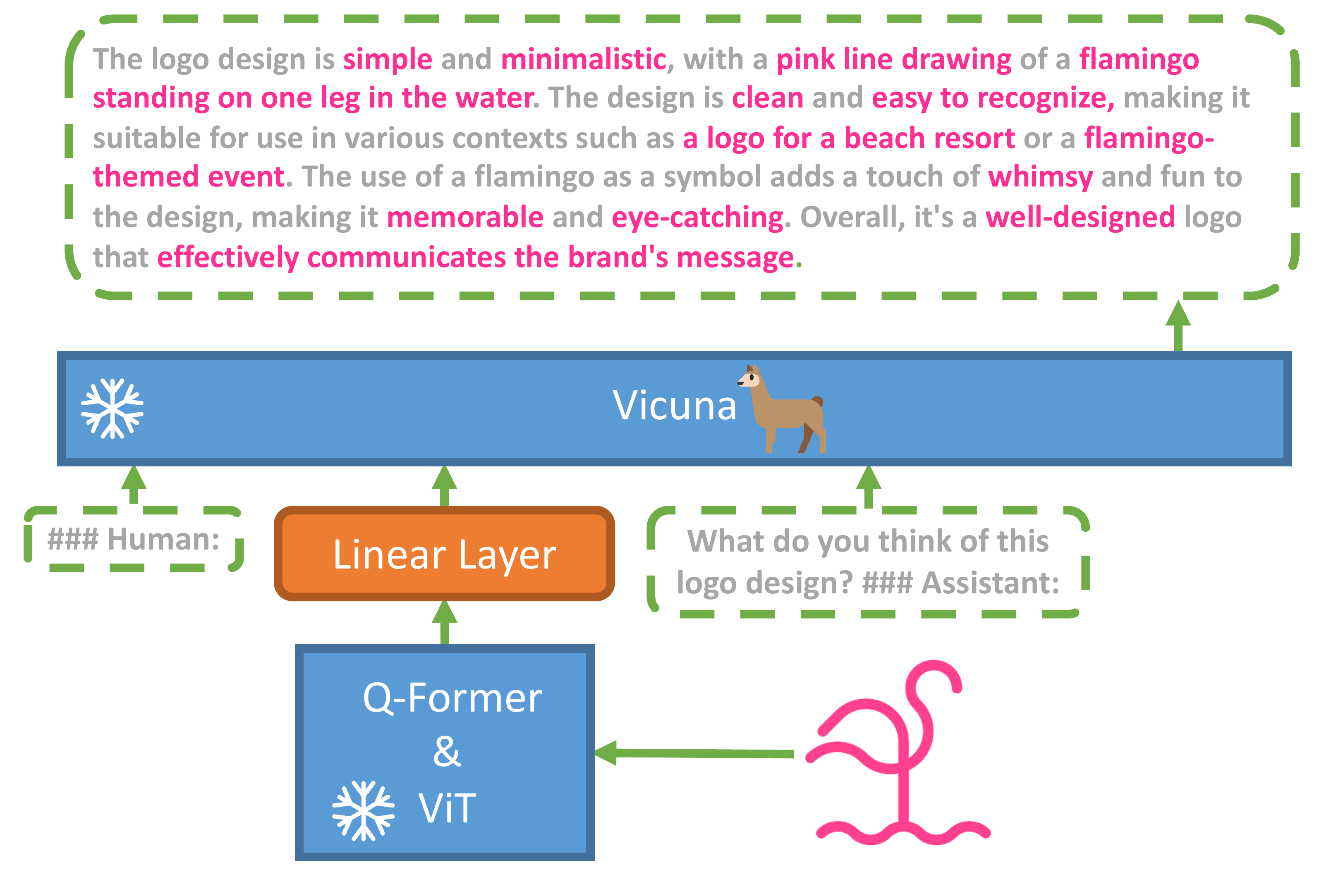}   
\caption{
\textbf{The architecture of MiniGPT-4.} It consists of a vision encoder with a pretrained ViT and Q-Former, a single linear projection layer, and an advanced Vicuna large language model. MiniGPT-4 only requires training the linear projection layer to align the visual features with the Vicuna. 
}
\label{fig:overview}
\vspace{-10pt}
\end{figure}

In our experiments, we discovered that MiniGPT-4 possesses numerous capabilities similar to those demonstrated by GPT-4. 
For instance, MiniGPT-4 can generate intricate image descriptions, create websites based on handwritten text instructions, and explain unusual visual phenomena. 
Furthermore, our findings revealed that MiniGPT-4 also has a variety of other intriguing abilities not showcased in the GPT-4 demonstrations. 
For example, MiniGPT-4 can directly generate detailed cooking recipes from food photos, write stories or poems inspired by images, write advertisements for products in images, identify problems shown in photos and provide corresponding solutions, and retrieve rich facts about people, movies, or art directly from images, among other capabilities. 
These abilities are absent in previous vision-language models like Kosmos-1 \citep{kosmos} and BLIP-2 \citep{blip2} that use less powerful language models. 
This further validates that integrating visual features with an advanced language model is one of the keys to enhancing vision-language models.

We present a summary of our key findings:

\begin{itemize}
    \item Our research reveals with compelling evidence that by aligning visual features with advanced large language models like Vicuna, MiniGPT-4 can achieve advanced vision-language capabilities comparable to those exhibited in the GPT-4 demonstrations.

    \item Our findings suggest that training merely one projection layer can effectively align a pretrained vision encoder with the large language model. Our MiniGPT-4 only requires training approximately 10 hours on 4 A100 GPUs. 

    \item We discovered that simply aligning visual features with large language models using short image caption pairs is not sufficient for developing a well-performing model and leads to unnatural language generation. Further finetuning with a small but detailed image description pairs can address this limitation and significantly improves its usability.
\end{itemize}

\section{Related Works}
\noindent \textbf{Large language models} have experienced tremendous success in recent years due to the scaling up of training data and an increase in the number of parameters. Early models, such as BERT~\citep{bert}, GPT-2~\citep{gpt2}, and T5~\citep{t5}, laid the foundation for this progress. Subsequently, GPT-3~\citep{gpt3}, with a massive scale of 175 billion parameters, was introduced, demonstrating significant breakthroughs across numerous language benchmarks. This development inspired the creation of various other large language models, including Megatron-Turing NLG~\citep{smith2022using}, Chinchilla~\citep{hoffmann2022training}, PaLM~\citep{chowdhery2022palm}, OPT~\citep{zhang2022opt}, BLOOM~\citep{scao2022bloom}, and LLaMA~\citep{llama}, among others. Wei \textit{et al.}~\citep{wei2022emergent} further discovered several \textit{emergent abilities}, which appear exclusively in large models. The emergence of these abilities underscores the importance of scaling up in the development of large language models. Moreover, by aligning the pre-trained large language model GPT-3 with human intent, instructions and human feedback, InstructGPT~\citep{instructGPT} and ChatGPT~\citep{chatgpt} enable conversational interactions with humans and can answer a wide range of diverse and complex questions. More recently, several open-sourced models,  such as Alpaca~\citep{alpaca} and Vicuna~\citep{vicuna2023}, have been developed based on LLaMA~\citep{llama} and also exhibit similar performance. 


\noindent \textbf{Leveraging Pre-trained LLMs in Vision-Language Tasks.}
In recent years, the trend of using autoregressive language models as decoders in vision-language tasks has gained significant traction~\citep{visualgpt,kosmos,yang2022zero,tiong2022plug,alayrac2022flamingo,blip2,blip1,palm_e}. This approach takes advantage of cross-modal transfer, allowing knowledge to be shared between language and multimodal domains. Pioneering studies like VisualGPT~\citep{visualgpt} and Frozen~\citep{tsimpoukelli2021multimodal} have demonstrated the benefits of employing a pre-trained language model as a vision-language model decoder. Flamingo~\citep{alayrac2022flamingo} was then developed to align a pre-trained vision encoder and language model using gated cross-attention, and was trained on billions of image-text pairs, showcasing impressive in-context few-shot learning capabilities. Following that, BLIP-2~\citep{blip2} was introduced, employing a Flan-T5~\citep{flanT5} with a Q-Former to efficiently align visual features with the language model. Most recently, PaLM-E~\citep{palm_e}, featuring 562 billion parameters, has been developed to integrate real-world continuous sensor modalities into an LLM, thereby establishing a connection between real-world perceptions and human languages. GPT-4~\citep{gpt4} has also been recently released, showcasing more powerful visual understanding and reasoning abilities after pre-training on a vast collection of aligned image-text data.

LLMs, such as ChatGPT, have proven to be powerful tools in enhancing the performance of vision-language tasks by collaborating with other specialized models. For instance, Visual ChatGPT~\citep{visualChatGPT} and MM-REACT~\citep{yang2023mmreact} showcase how ChatGPT can act as a coordinator, integrating with diverse visual foundation models and facilitating their collaboration to tackle more complex challenges. ChatCaptioner~\citep{chatcaptioner} treats ChatGPT as a questioner, prompting diverse questions for BLIP-2 to answer. Through multi-round conversations, ChatGPT extracts visual information from BLIP-2 and effectively summarizes the image content. Video ChatCaptioner~\citep{chen2023video} extends this approach, applying it to video spatiotemporal understanding. ViperGPT~\citep{vipergpt} demonstrates the potential of combining an LLM with different vision models to address complex visual queries programmatically. In contrast, MiniGPT-4 directly aligns visual information with the language model to accomplish diverse vision-language tasks without the usage of external vision models.

\section{Method}

MiniGPT-4 aims to align visual information from a pretrained vision encoder with an advanced large language model (LLM). Specifically, we utilize the Vicuna~\citep{vicuna2023} as our language decoder, which is constructed upon LLaMA~\citep{llama} and can perform a wide range of complex linguistic tasks. For visual perception, we employ the same visual encoder as used in BLIP-2~\citep{blip2}, a ViT backbone~\citep{fang2022eva} coupled with their pre-trained Q-Former. Both language and vision models are open-sourced. We target to bridge the gap between the visual encoder and LLM using a linear projection layer, with an overview of our model displayed in Fig.\ref{fig:overview}.

To achieve an effective MiniGPT-4, we propose a two-stage training approach. The initial stage involves pretraining the model on a large collection of aligned image-text pairs to acquire vision-language knowledge. In the second stage, we finetune the pretrained model with a smaller but high-quality image-text dataset with a designed conversational template to enhance generation reliability and usability.

\subsection{First pretraining stage}

During the initial pretraining stage, the model is designed to acquire vision-language knowledge from a large collection of aligned image-text pairs. We regard the output from the injected projection layer as a soft prompt for the LLM, prompting it to generate the corresponding ground-truth texts.

Throughout the entire pretraining process, both the pretrained vision encoder and the LLM remain frozen, with only the linear projection layer being pretrained. We use a combined dataset of Conceptual Caption \citep{changpinyo2021conceptual,sharma2018conceptual},  SBU \citep{ordonez2011im2text} and LAION \citep{laion} to train our model. Our model undergoes 20,000 training steps with a batch size of 256, covering approximately 5 million image-text pairs. The entire process takes about 10 hours to complete, utilizing 4 A100 (80GB) GPUs.

\textbf{Issues of the first pretraining stage} 
Following the first pretraining stage, our MiniGPT-4 demonstrates the capacity to possess a wealth of knowledge and offer reasonable responses to human inquiries. However, we have observed instances where it produces incoherent linguistic outputs, such as repetitive words or sentences, fragmented sentences, or irrelevant content. These issues hinder MiniGPT-4's ability to engage in a fluent visual conversation with humans.

We also observed similar challenges encountered in GPT-3. Despite its pretraining on a extensive language dataset, GPT-3 struggles to generate language outputs that are accurately aligned with users' intentions.
Through a process of instruction fine-tuning and reinforcement learning from human feedback, GPT-3 evolves into GPT-3.5 \citep{instructGPT, chatgpt} and becomes capable of producing more human-friendly outputs. 
This phenomenon bears a resemblance to the current state of MiniGPT-4 following its initial pretraining stage. As such, it is not surprising that our model may struggle to generate fluent and natural human language outputs at this stage.

\subsection{Curating a high-quality alignment dataset for vision-language domain.}
To achieve greater naturalness in the generated language and enhance the model's usability, a second-stage alignment process is essential. While in the realm of NLP, instruction fine-tuning datasets \citep{alpaca} and conversations \citep{sharegpt} are easily accessible, no equivalent datasets exist for the vision-language domain. To address this deficiency, we carefully curated a detailed image description dataset, specifically tailored for vision-language alignment purposes. This dataset is subsequently utilized to fine-tune our MiniGPT-4 during the second-stage alignment process.

\paragraph{Initial aligned image-text generation}
In the initial phase, we employ the model derived from the first pretraining stage to generate comprehensive descriptions of input images. To enable our model to produce more detailed image descriptions, we designed a prompt that adheres to the conversational format of the Vicuna \citep{vicuna2023} language model, as shown below. In this prompt, \textit{\textless ImageFeature\textgreater } represents the visual features produced by the linear projection layer.

\textit{\#\#\#Human: \textless Img\textgreater \textless ImageFeature\textgreater \textless /Img\textgreater  Describe this image in detail. Give as many details as possible. Say everything you see. \#\#\#Assistant: }

To identify incomplete sentences, we examine whether the generated sentence exceeds 80 tokens.  If it does not, we incorporate an additional prompt,  \textit{\#\#\#Human: Continue \#\#\#Assistant: }, prompting our MiniGPT-4 to extend the generation process. By concatenating the outputs from both steps, we can create a more comprehensive image description. This approach enables us to generate image-text pairs with detailed and informative image descriptions. We randomly select 5,000 images from the Conceptual Caption dataset \citep{changpinyo2021conceptual,sharma2018conceptual} and use the pretrained model to generate corresponding language descriptions for each image.

\paragraph{Data post-processing}
The above automatically generated image descriptions contain noisy or incoherent descriptions, such as repetition of words or sentences, fragmented sentences, or irrelevant content. 
In order to fix these issues, we employ ChatGPT to mend the descriptions by utilizing the following prompt:

\textit{Fix the error in the given paragraph. Remove any repeating sentences, meaningless characters, not English sentences, and so on. Remove unnecessary repetition. Rewrite any incomplete sentences. Return directly the results without explanation. Return directly the input paragraph if it is already correct without explanation.}

Upon completing the post-processing stage, we manually verify the correctness of each image description to guarantee its high quality. Specifically, we first identified several frequently shown errors (\textit{``I'm sorry I made a mistake...'', or ``I apologize for that ...''}) and then hard-coded rules to automatically filter them out. We also manually refine the generated captions by eliminating redundant words or sentences that ChatGPT fails to detect. Finally, only approximately 3,500 out of 5,000 image-text pairs satisfy our requirement, and these pairs are subsequently utilized for the second-stage alignment process.

\subsection{Second-stage finetuning}
During the second stage, we finetune our pretrained model with the curated high-quality image-text pairs. During the finetuning, we use the predefined prompts in the following template:

\textit{\#\#\#Human: \textless Img\textgreater \textless ImageFeature\textgreater \textless /Img\textgreater  \textless Instruction\textgreater  \#\#\#Assistant:}

In this prompt, \textit{\textless Instruction\textgreater } represents a randomly sampled instruction from our predefined instruction set containing variant forms of instructions such as \textit{``Describe this image in detail''} or \textit{``Could you describe the contents of this image for me''}. 
It is important to note that we do not calculate the regression loss for this specific text-image prompt.

As a result, MiniGPT-4 is now capable of producing more natural and reliable language outputs. Furthermore, we observed that this fine-tuning process is remarkably efficient, only requiring a mere 400 training steps with a batch size of 12, which takes around 7 minutes with a single A100 GPU.

\section{Experiments}

In the experiment, we aim to showcase the diverse and emergent capabilities of our MiniGPT-4 model through various qualitative examples. These abilities include generating detailed image descriptions, identifying amusing aspects within memes, providing food recipes from photos, writing poems for images, etc. Additionally, we present quantitative results on the task of image captioning. 


\begin{figure}[t]
\begin{minipage}{0.5\textwidth}
\centering
 \includegraphics[width=0.9\linewidth]
    {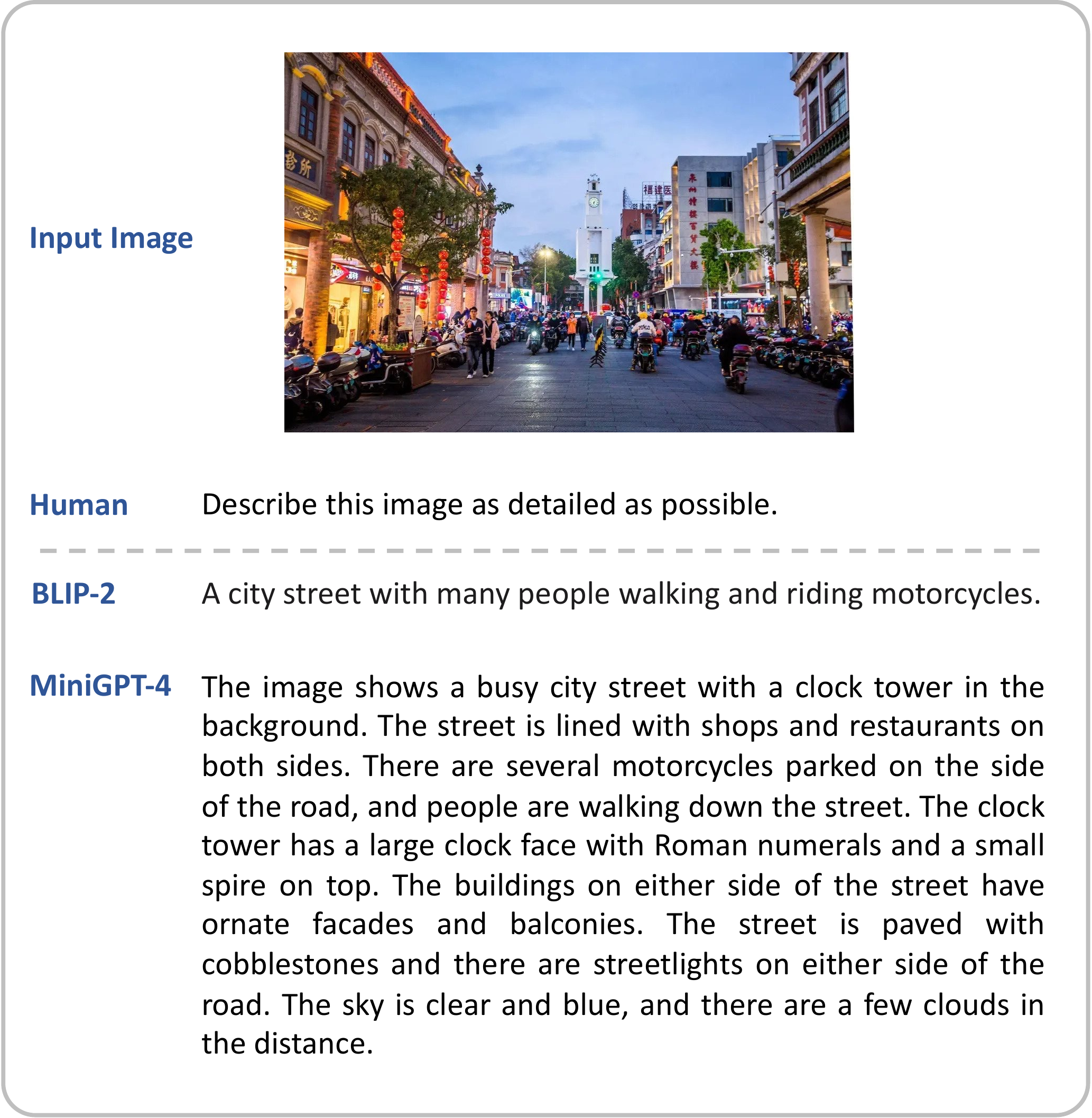}   
\caption{Detailed description}
\label{fig:detailed}
\end{minipage} \hfill
\begin{minipage}{0.5\textwidth}
\centering
 \includegraphics[width=0.9\linewidth]
    {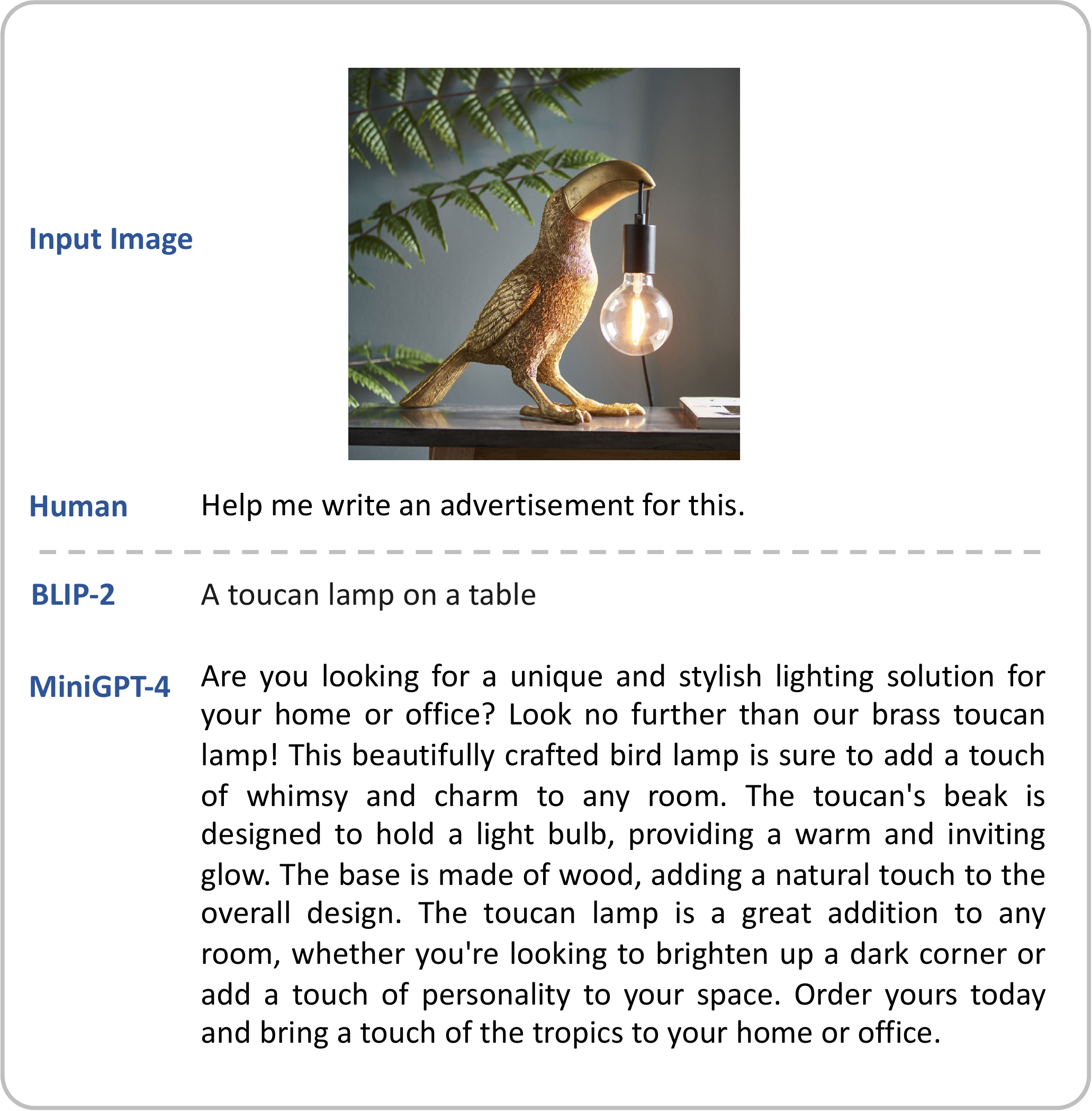}   
\caption{Advertisement promotion}
\label{fig:ad}
\end{minipage} \hfill
\vspace{-10pt}
\end{figure}

\subsection{Uncovering emergent abilities with MiniGPT-4 through qualitative examples} 

MiniGPT-4 demonstrates many advanced abilities compared to traditional vision-language models. For example, it can describe images in detail and interpret the humorous aspects of a given meme. Here, we qualitatively compared our model to one of the leading vision-language models, BLIP-2 \citep{blip2}, with eight distinct examples, each highlighting a different ability.

\begin{figure}[t]
  \begin{subfigure}{0.55\textwidth}
    \centering
    \includegraphics[width=0.9\linewidth]{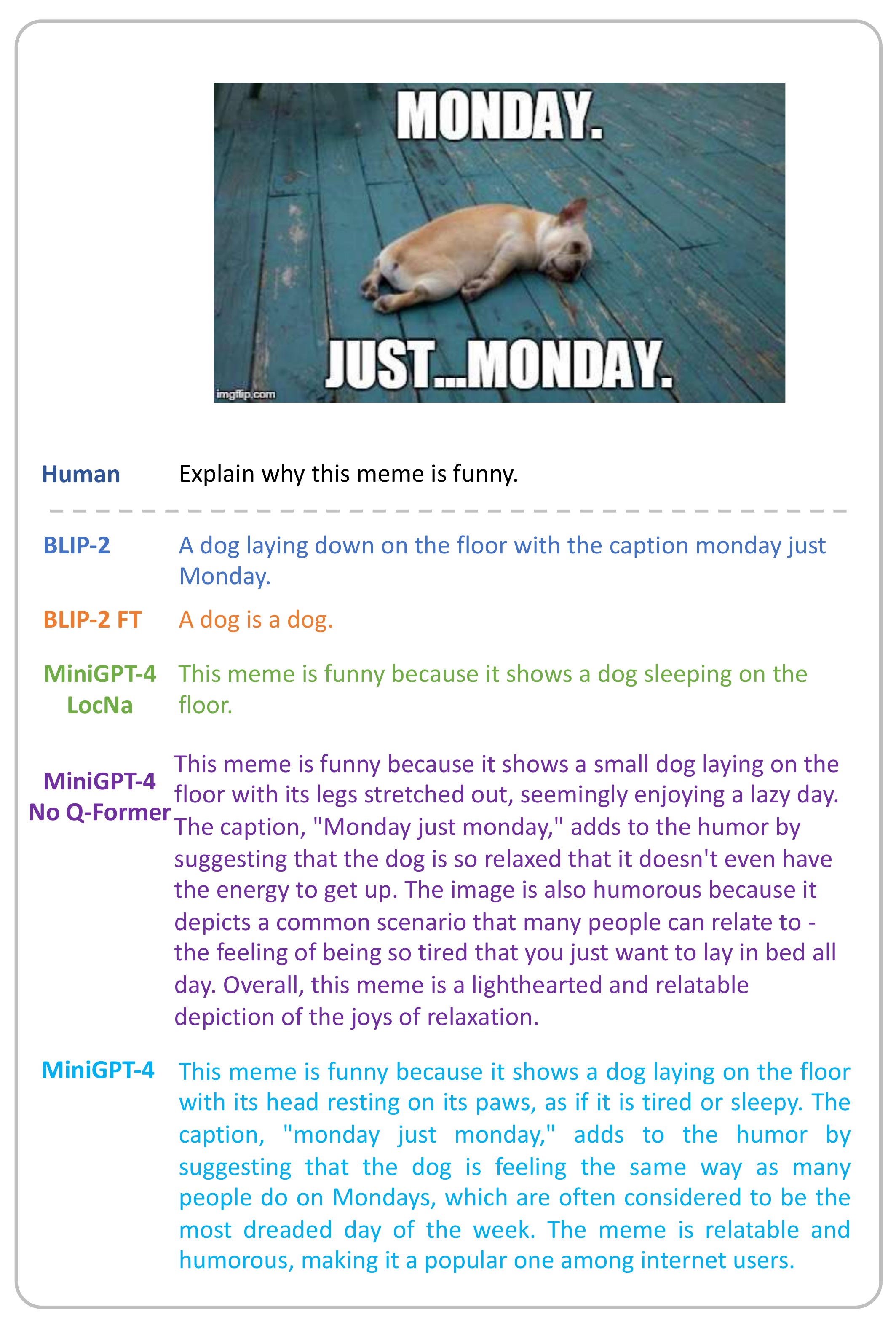}
    \caption{Meme explaining}
    \label{fig:ab_meme}
  \end{subfigure}%
  \hfill
  \begin{subfigure}{0.45\textwidth}
    \centering
    \includegraphics[width=0.9\linewidth]{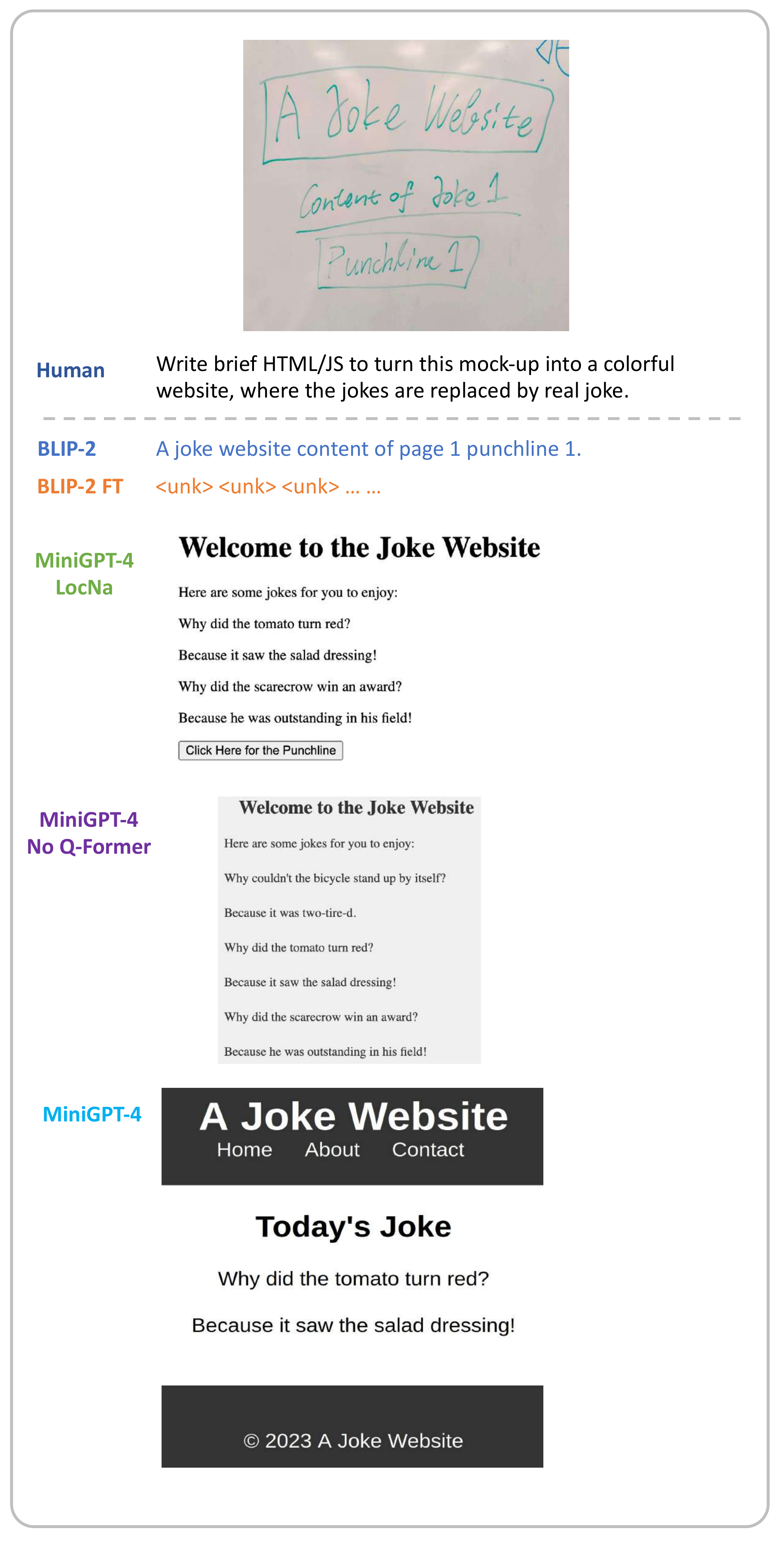}
    \caption{Website Creating}
    \label{fig:ab_website}
  \end{subfigure}
  \caption{Model generations from BLIP-2, BLIP-2 finetuned our second stage data (BLIP-2 FT), MiniGPT-4 finetuned with Local Narrative data in the second stage (MiniGPT-4 LocNa), MiniGPT-4 model without Q-Former (MiniGPT-4 No Q-Former), and MiniGPT-4.}
  \label{fig:overall_label}
  \vspace{-10pt}
\end{figure}

An example in Fig.\ref{fig:detailed} demonstrates that MiniGPT-4 effectively identifies various elements within the image, such as busy city streets, clock towers, shops, restaurants, motorcycles, people, streetlights, and clouds. In contrast, BLIP-2 can only cover city streets, people, and motorcycles in its image caption generation. Another example presented in Fig.\ref{fig:ab_meme} shows that MiniGPT-4 successfully explains why the meme is humorous. It interprets that the lying dog is feeling the same way as many people do on Monday, which is often considered to be the most dreaded day of the week. In contrast, BLIP-2 only briefly describes the image content and fails to comprehend the amusing aspects of the image.

We also showcase MiniGPT-4's other abilities by demonstrating other distinctive abilities. These include creating advertising promotions based on a given image (Fig.\ref{fig:ad}), retrieving factual information from a movie photograph (Fig.\ref{fig:movie}), generating a food recipe from a food image (Fig.\ref{fig:cook}), diagnosing plant diseases and suggesting treatment plans (Fig.\ref{fig:plant}), creating a website from a hand-written draft (Fig.\ref{fig:ab_website}), and writing poems inspired by an image (Fig.\ref{fig:poem}). These abilities are absent in traditional vision-language models like BLIP-2 (utilizing Flan-T5 XXL~\citep{flanT5} as a language model), which use less powerful language models (LLMs). This contrast indicates that those advanced vision-language abilities only emerge when the visual features are properly aligned with an advanced LLM such as Vicuna \citep{vicuna2023}.

\begin{table}[b]
\vspace{2mm}
\centering
\caption{Quantitative results on advanced vision-language tasks. MiniGPT-4 shows strong performance and successfully responses to 65\% of the requests.}
\scalebox{0.9}{
\begin{tabular}{lccccc}
\toprule
 & Meme & Recipes & Ads & Poem & Avg. \\
\midrule
BLIP-2 & 0/25 & 4/25 & 1/25 & 0/25 & 5/100 \\
MiniGPT-4 & 8/25 & 18/25 & 19/25 & 20/25 & 65/100 \\ 
\bottomrule
\label{tab: quanti_adv}
\end{tabular}
}
\end{table}

\subsection{Quantitative analysis}

\begin{table}[b]
\begin{minipage}{0.45\textwidth}
\vspace{2mm}
\centering
\caption{COCO caption evaluation. We use ChatGPT to judge if the generated caption covers all the visual objects and relations in the ground-truth caption.}
\scalebox{0.75}{
\begin{tabular}{ccc}
\toprule
 & BLIP-2 & MiniGPT-4 \\
\midrule
Correctness & 1376/5000 & 3310/5000  \\
Percentage & 27.5\% & 66.2\% \\ 
\bottomrule
\label{human_evaluation}
\end{tabular}
}
\end{minipage}
\hfill
\begin{minipage}{0.45\textwidth}
\centering
\caption{Failure rates of detailed caption and poem generation tasks before and after second-stage finetuning. The finetuning stage significantly  reduces generation failures.}
\scalebox{0.75}{
\begin{tabular}{ccc}
\toprule
 Failure rate &   Detailed caption   & Poem\\
\midrule
 Before stage-2 & 35\% & 32\% \\ 
 After stage-2 & 2\% & 1\%  \\
\bottomrule
\end{tabular}
\label{exp:stage2ablation}
}
\end{minipage}
\end{table}

\begin{figure}[t]
\begin{minipage}{0.485\textwidth}
\centering
 \includegraphics[width=0.9\linewidth]
    {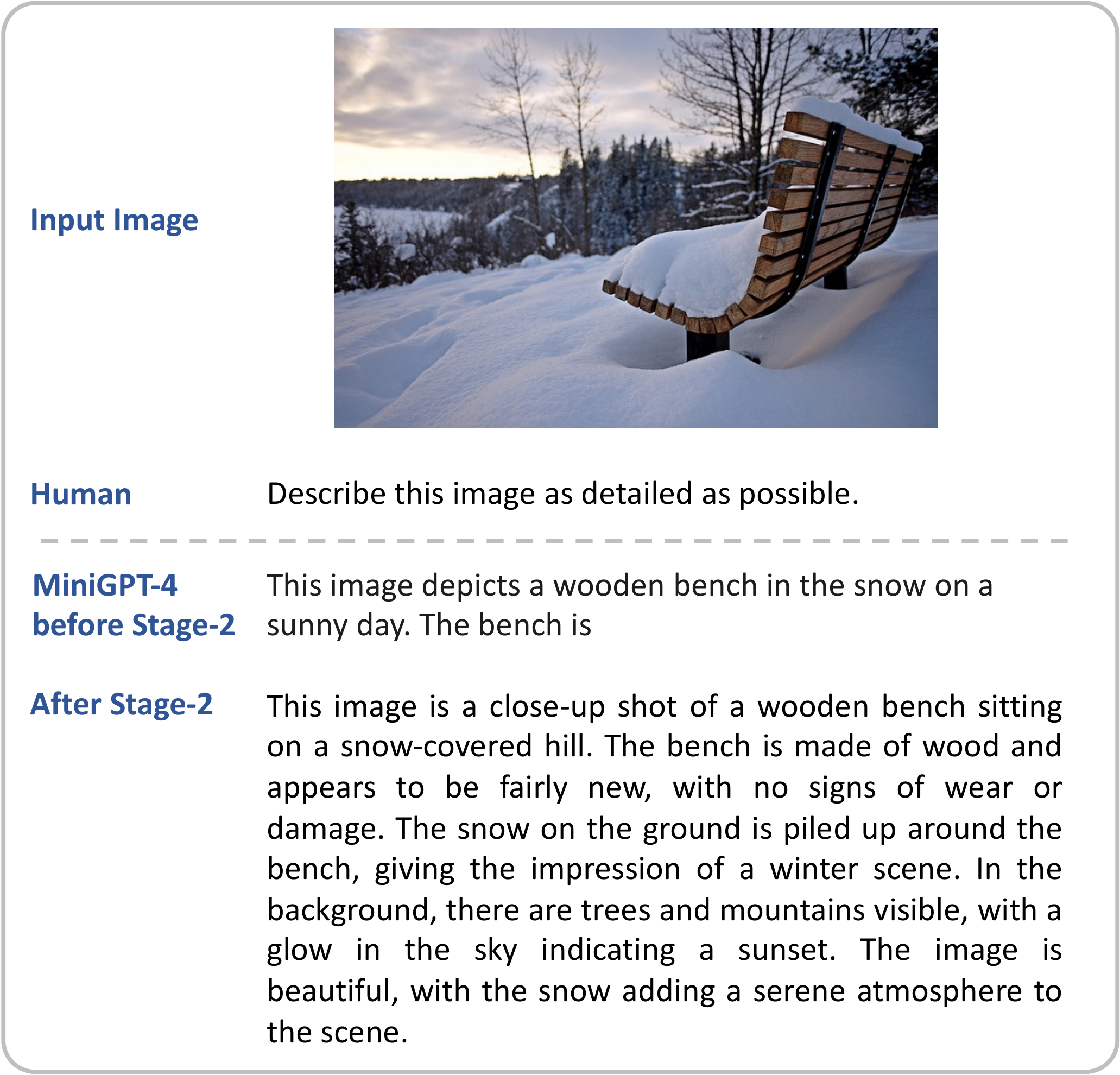}   
\caption{MiniGPT-4 before second-stage finetuning fails to output completed texts. The generation is improved after the finetuning.}
\label{fig:secondstage}
\end{minipage} \hfill
\begin{minipage}{0.485\textwidth}
\centering
 \includegraphics[width=0.9\linewidth]
    {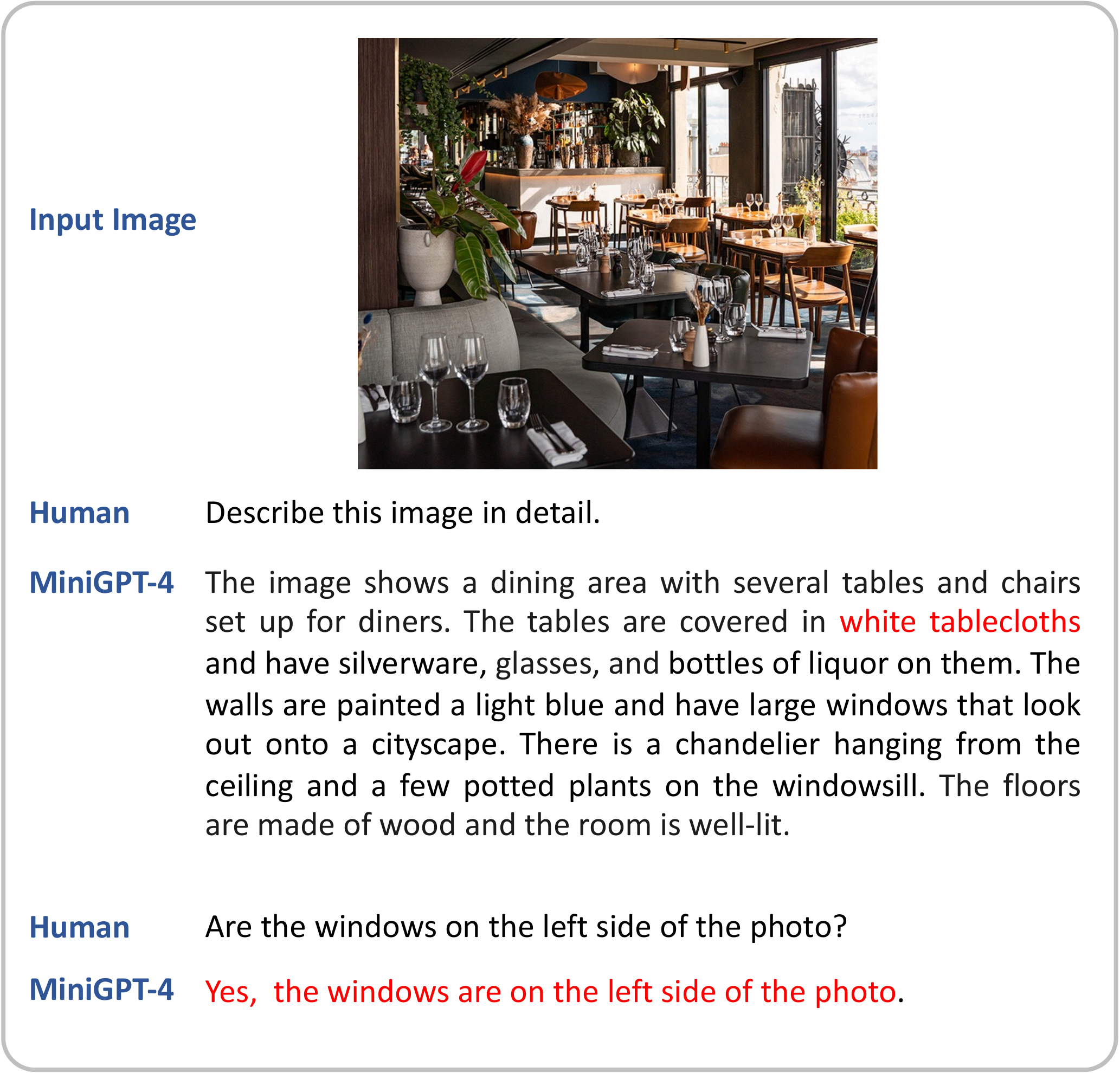}  
\caption{An example of MiniGPT-4's limitations. MiniGPT-4 hallucinates unexisting tablecloths and can't locate the windows correctly.}
\label{fig:Limitation}
\end{minipage} \hfill
\vspace{-10pt}
\end{figure}

\paragraph{Advanced Abilities}
To quantify performance on advanced vision-language tasks, we compiled a small evaluation dataset comprising 4 tasks: meme interpretation with the question ``Explain why this meme is funny.'', recipe generation with the question ``How should I make something like this?'', advertisement creation with the prompt ``Help me draft a professional advertisement for this.'', and poem composition with ``Can you craft a beautiful poem about this image?''.
In total, we collect 100 diverse images, with 25 images allocated to each task.
We asked human evaluators to determine whether the model generation satisfies the request.
We compared our results with BLIP-2 \citep{blip2} and present the findings in Tab.\ref{tab: quanti_adv}. In meme interpretation, poem writing, and advertisement creation, BLIP-2 largely struggles to fulfill any requests. For recipe generation, BLIP-2 succeeds in 4 out of 25 cases. In contrast, MiniGPT-4 manages to address the requests in recipes, advertisements, and poem generation in nearly 80\% of the instances. Furthermore, MiniGPT-4 correctly comprehends the challenging humor understanding in memes in 8 out of 25 cases.

\paragraph{Image Captioning} We evaluate the performance of MiniGPT-4 on the COCO caption benchmark and compare it with BLIP-2 \citep{blip2}. 
Our model's generated captions typically contain rich visual details. As such, conventional similarity-based image-caption evaluation metrics struggle to provide an accurate evaluation of our models. 
In this regard, we evaluate the performance by checking if the generated captions cover all the ground truth captions' information with the help of ChatGPT and details can be found in Appx.\ref{appx: caption_eval}. Results in Tab.\ref{human_evaluation} shows that MiniGPT-4 outperforms BLIP-2 in generating captions that are more closely aligned with the ground-truth visual objects and relationships. With a success rate of 66.2\%, MiniGPT-4 is considerably more accurate than BLIP-2, which achieves only 27.5\%. 
Further evaluation on traditional VQA tasks can be found in Appx.\ref{appx: vqa}.

\subsection{Analysis on the second-stage finetuning}

\paragraph{Effectiveness of the second-stage finetuning}

The utilization of only the model pretrained after the first pretraining stage may result in failures, such as the occurrence of repetitive words or sentences, fragmented sentences, or irrelevant content. However, these issues have been largely mitigated through the second-stage finetuning process. This can be observed in Fig.\ref{fig:secondstage}, where MiniGPT-4 generates incomplete captions before the second-stage finetuning. However, after the second-stage finetuning, MiniGPT-4 is capable of generating complete and fluent captions. In this section, we investigate the importance and effectiveness of the second-stage finetuning approach.

To quantify the impact of second-stage finetuning, we randomly sampled 100 images from the COCO test set and investigated the model performance on two tasks: detailed description generation and poem writing. The prompts used were ``\textit{Describe the image in detail.}'' and ``\textit{Can you write a beautiful poem about this image?}''. These tasks were performed by both the models before and after second-stage finetuning. We manually counted the number of failure generations for the model in each stage. The results are presented in Tab.\ref{exp:stage2ablation}. Prior to the second-stage finetuning, approximately 1/3 of the generated outputs failed to match ground truth captions or poems. In contrast, the model after second-stage fineuning has less than two failure cases out of the 100 test images for both tasks. These experimental results demonstrate that second-stage finetuning yields a significant improvement in the quality of generated outputs. A qualitative example of the model generation before and after the second-stage finetuning is shown in Fig.\ref{fig:secondstage}.

\paragraph{Can the original BLIP-2 benefit from the second-stage data?} 
In this study, we finetune BLIP-2 \citep{blip2} with our second-stage data in the same way as MiniGPT-4, and check if it can obtain similar advanced abilities as MiniGPT-4. The finetuned BLIP-2 is denoted as BLIP-2 FT. 
Note that MiniGPT-4 uses the same visual module as BLIP-2; while BLIP-2 uses FlanT5 XXL \citep{flanT5} as the language model, which is not as strong as the Vicuna \citep{vicuna2023} model used in our MiniGPT-4 model.
We rely on the same prompts to assess the advanced capabilities of our model. 
Qualitative results are shown in Fig.\ref{fig:overall_label}, \ref{fig:ab_cook}, and \ref{fig:ab_des}. We discover that BLIP-2 FT still generates short responses and fails to generalize to advanced tasks like meme explaining and website coding (Fig.\ref{fig:overall_label}). Our finding suggests that BLIP-2's relatively weaker language model FlanT5 XXL benefits less from such a small dataset, and highlights the effectiveness of a more advanced LLM in a VLM system.

\begin{table}[b]
\begin{minipage}{0.45\textwidth}
\centering
\caption{Ablation on architecture designs}
\scalebox{0.75}{
\begin{tabular}{lcc}
    \toprule
    \textbf{Model} & \textbf{AOK-VQA} & \textbf{GQA} \\
    \midrule
    MiniGPT-4 & 58.2 & 32.2 \\
    (a) MiniGPT-4 w/o Q-Former & 56.9 & 33.4 \\
    (b) MiniGPT-4 + 3 Layers & 49.7 & 31.0 \\
    (c) MiniGPT-4 + Finetune Q-Former & 52.1 & 28.0 \\
    \bottomrule
\end{tabular}
\label{tab: ablation}
}
\end{minipage} \hfill
\begin{minipage}{0.45\textwidth}
\centering
\caption{Hallucination Evaluation}
\scalebox{0.75}{
\begin{tabular}{lcc}
\toprule
                    & $\textbf{CHAIR}_i$  & \textbf{Avg. Length} \\
\midrule
Blip-2              & 1.3         & 6.5        \\
MiniGPT-4 (short)   & 7.2         & 28.8       \\
MiniGPT-4 (long)    & 9.6         & 175        \\
\bottomrule
\end{tabular}
\label{tab:hallu}
}
\end{minipage}
\end{table}

\paragraph{Second stage with Localized Narratives }
The dataset Localized Narratives \citep{pont2020connecting} is a detailed image description dataset where 
annotators describe images while simultaneously localizing the corresponding regions.
Here, we test the performance of our model by replacing our self-collected dataset in the second-stage with the Localized Narratives dataset. The model is denoted as MiniGPT-4 LocNa. Qualitative results in Fig.\ref{fig:overall_label}, \ref{fig:ab_cook}, and \ref{fig:ab_des} show that MiniGPT-4 LocNa can generate long image descriptions (Fig.\ref{fig:ab_des}). However, the generated outputs have lower quality with monotonous expressions. Besides, MiniGPT-4 LocNa does not generalize as well as the original MiniGPT-4 in other complex tasks like explaining why the meme is funny (Fig.\ref{fig:ab_meme}). The performance gap may be due to the monotonous and repeated image descriptions in Localized Narratives.

\subsection{Ablation on the architecture designs}
To further demonstrate the effectiveness of using one single linear layer to align visual features with LLM, we conduct experiments with different architecture designs, including (a) removing the Q-Former and directly mapping the VIT’s output to Vicuna’s embedding space (i.e., without Q-former), (b) using three linear layers instead of one layer, and (c) additionally finetuning the Q-Former in the vision module. 
All the variants are trained in the same way as the original design. Results on AOK-VQA \citep{schwenk2022okvqa} and GQA \citep{hudson2019gqa} datasets in Tab.\ref{tab: ablation} show that the variant (a) \textbf{MiniGPT-4 w/o Q-Former} has a similar performance to the original design. 
Qualitative results of this variant in Fig.\ref{fig:overall_label}, \ref{fig:ab_cook}, and \ref{fig:ab_des} also show similar advanced skills.
This reveals that the Q-Former from BLIP-2 doesn't plays a critical roles for advanced skills. 
Besides, both variants (b) \textbf{MiniGPT-4+ 3 Layers} and (c) \textbf{MiniGPT-4 + finetuning Q-Former}, perform slightly worse than the original MiniGPT-4. 
This indicates a single projection layer is sufficient to align the vision encoder and the large language model in our limited training data setting.

\subsection{Limitation analysis}
\paragraph{Hallucination}
As MiniGPT-4 is built upon LLMs, it inherits LLM's limitations like hallucinating nonexistent knowledge. An example in Fig. \ref{fig:Limitation} shows that MiniGPT-4 incorrectly identifies the presence of white tablecloths in the image, despite their absence. Here, we use the metric $\text{CHAIR}_i$ \citep{rohrbach2018object} to gauge the hallucination rate of the generation, with the two distinct prompts to control the model generation length: \textit{MiniGPT-4 (long)}: Please describe this image as detailed as possible. \textit{MiniGPT-4 (short)}: Please describe the image shortly and precisely, in less than 20 words.

Results in Tab.\ref{tab:hallu} show that longer captions tend to have higher hallucination rates. For example, MiniGPT-4 (long) generates captions averaging 175 words with a higher hallucination rate, while MiniGPT-4 (short) averages 28.8 words with a lower rate. BLIP-2, averaging 6.5 words, hallucinates less but covers fewer objects as seen in Tab.\ref{human_evaluation}.
Hallucination in detailed image descriptions is still an unresolved issue. Using Reinforcement Learning with AI feadback with hallucination detection modules may be a potential solution.

\paragraph{Spatial Information Understanding} MiniGPT-4's visual perception remains limited. It may struggle to differentiate spatial localization. For example, MiniGPT-4 in Fig. \ref{fig:Limitation} fails to identify the location of the windows. 
This limitation may stem from a lack of aligned image-text data designed for spatial information understanding. Training on such datasets like RefCOCO \citep{kazemzadeh2014referitgame} or Visual Genome \citep{krishna2017visual} could potentially alleviate this issue.

\section{Discussion}
How does MiniGPT-4 obtain these advanced abilities?
Many of the advanced vision-language capabilities demonstrated by GPT-4 can be understood as compositional skills rooted in two foundational skills: image understanding and language generation.
Take the task of image-based poem writing as an example. Advanced LLMs like ChatGPT and Vicuna can already craft poems based on users' instructions.
If they acquire the ability to understand images, compositionally generalizing to the task of image-based poem writing even without having image-poem pairs in their training data is possible.

In the first pretraining stage, MiniGPT-4 learns to understand images by modeling the correlation between images and short image descriptions from image caption datasets.
However, the language style in these image caption datasets differs from that of modern LLMs' generation, which leads to distorted language generation and hinders successful compositional generalization.
Therefore, we introduce a second-stage finetuning to restore the language generation ability.
MiniGPT-4 after the two-stage training successfully generalizes to many advanced compositional vision-language abilities like website coding from drafts or meme interpretation, verifies our assumption.
Future research might delve deeper into the mechanism of compositional generalization and seek ways to enhance them. We hope our work, as an early exploration of these vision-based LLM capabilities, will spur further investigations in this domain.

\bibliography{iclr2024_conference}
\bibliographystyle{iclr2024_conference}

\newpage
\appendix
\section{Appendix}
\subsection{More Qualitative Results}

\begin{figure}[ht]
\begin{minipage}{0.5\textwidth}
\centering
 \includegraphics[width=1.0\linewidth]
    {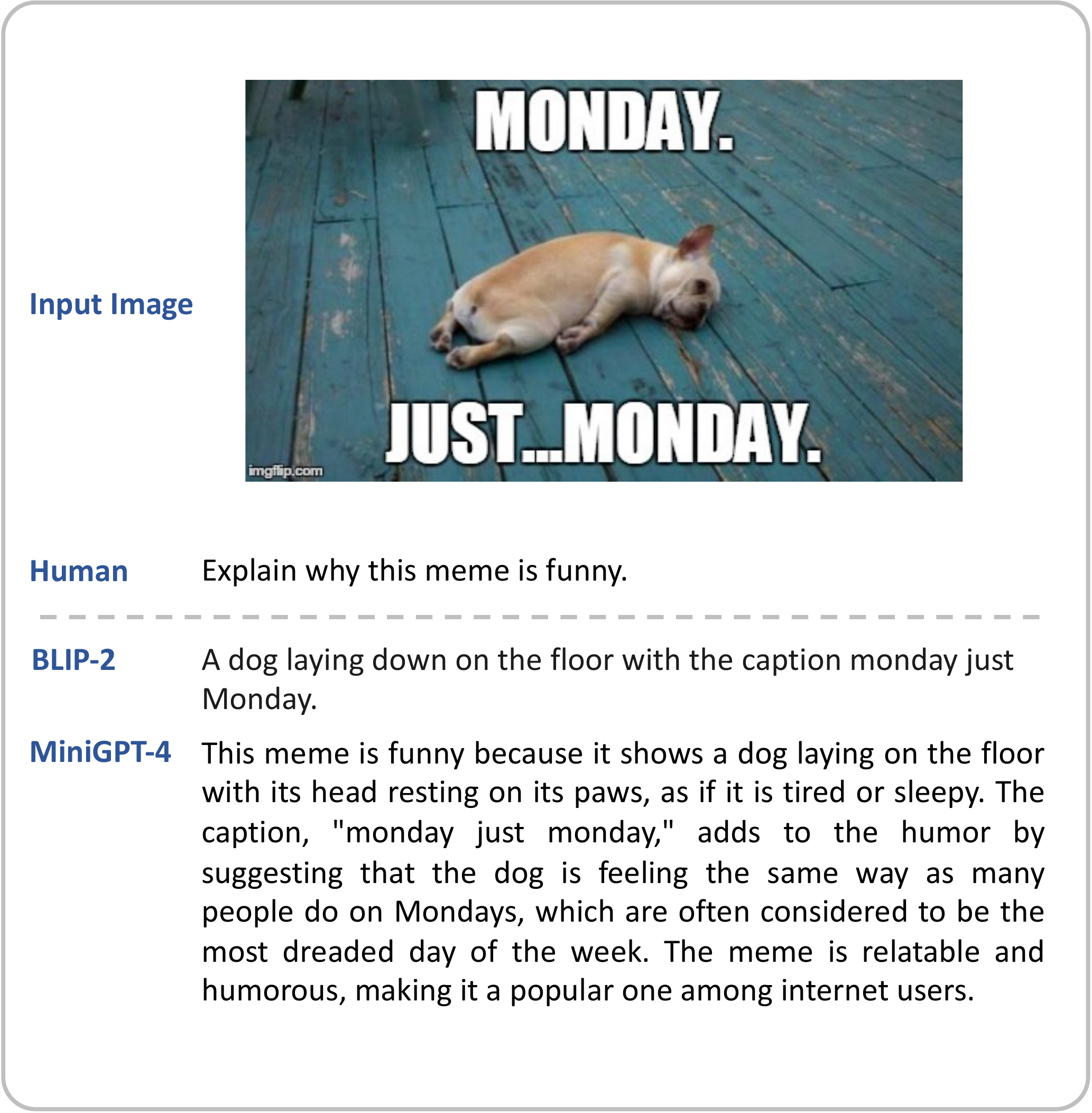}   
\caption{Advertisement promotion}
\label{fig:meme}
\end{minipage} \hfill
\begin{minipage}{0.5\textwidth}
\centering
 \includegraphics[width=1.0\linewidth]
    {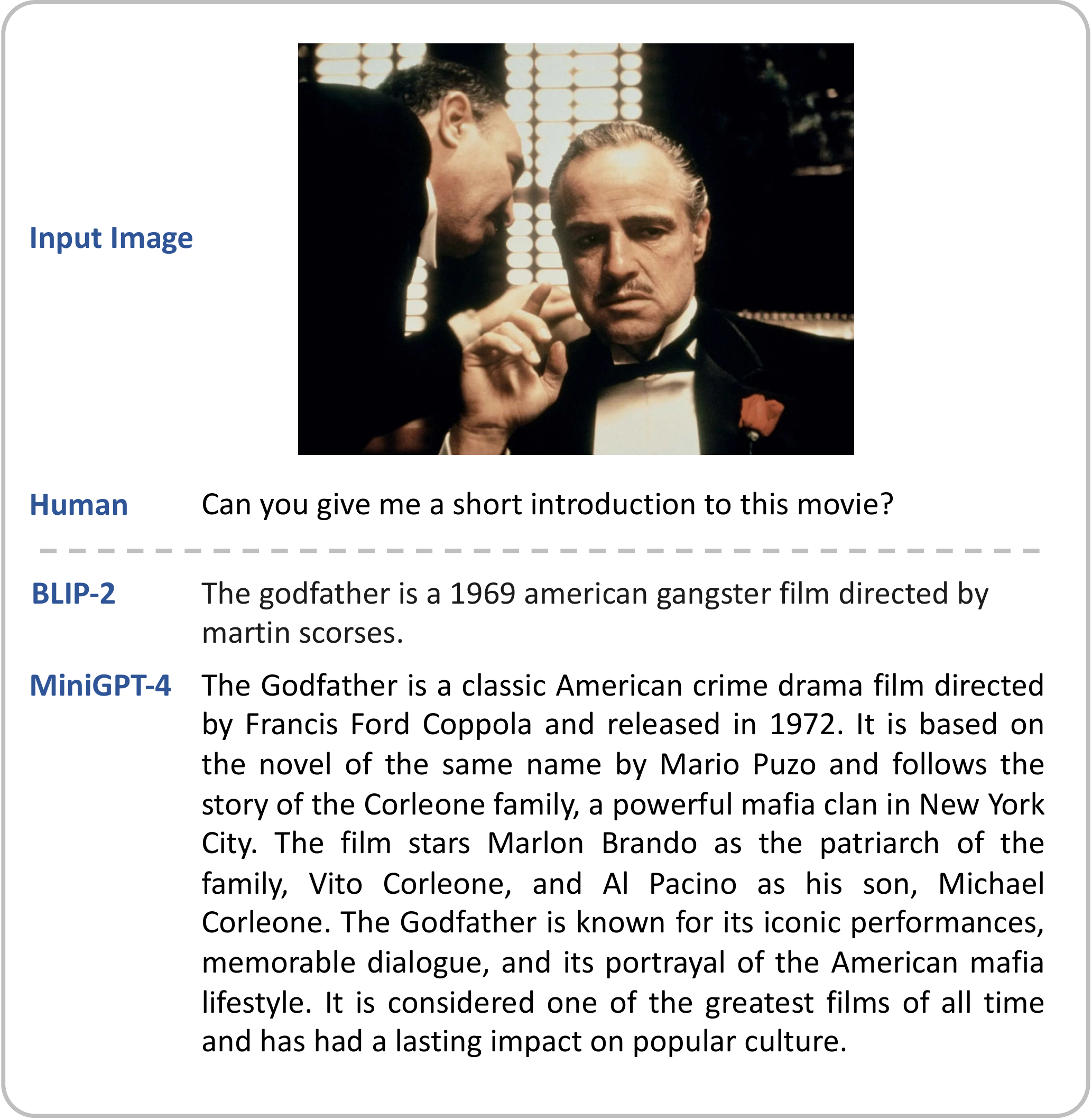}  
\caption{Factual retrieval}
\label{fig:movie}
\end{minipage} \hfill
\end{figure}

\begin{figure}[ht]
\begin{minipage}{0.5\textwidth}
\centering
 \includegraphics[width=1.0\linewidth]
    {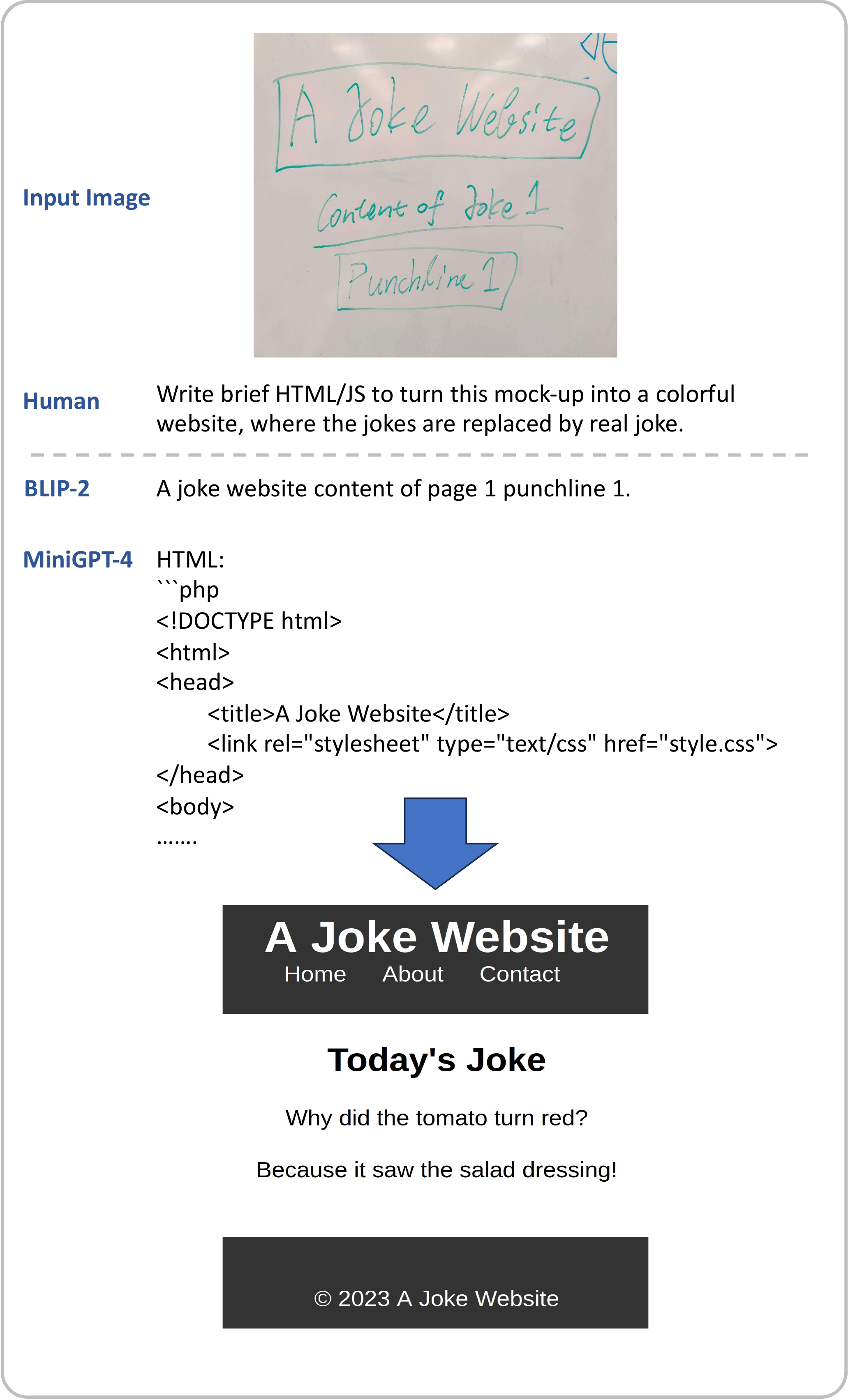}   
\caption{Website creating}
\label{fig:website}
\end{minipage} \hfill
\begin{minipage}{0.5\textwidth}
\centering
 \includegraphics[width=1.0\linewidth]
    {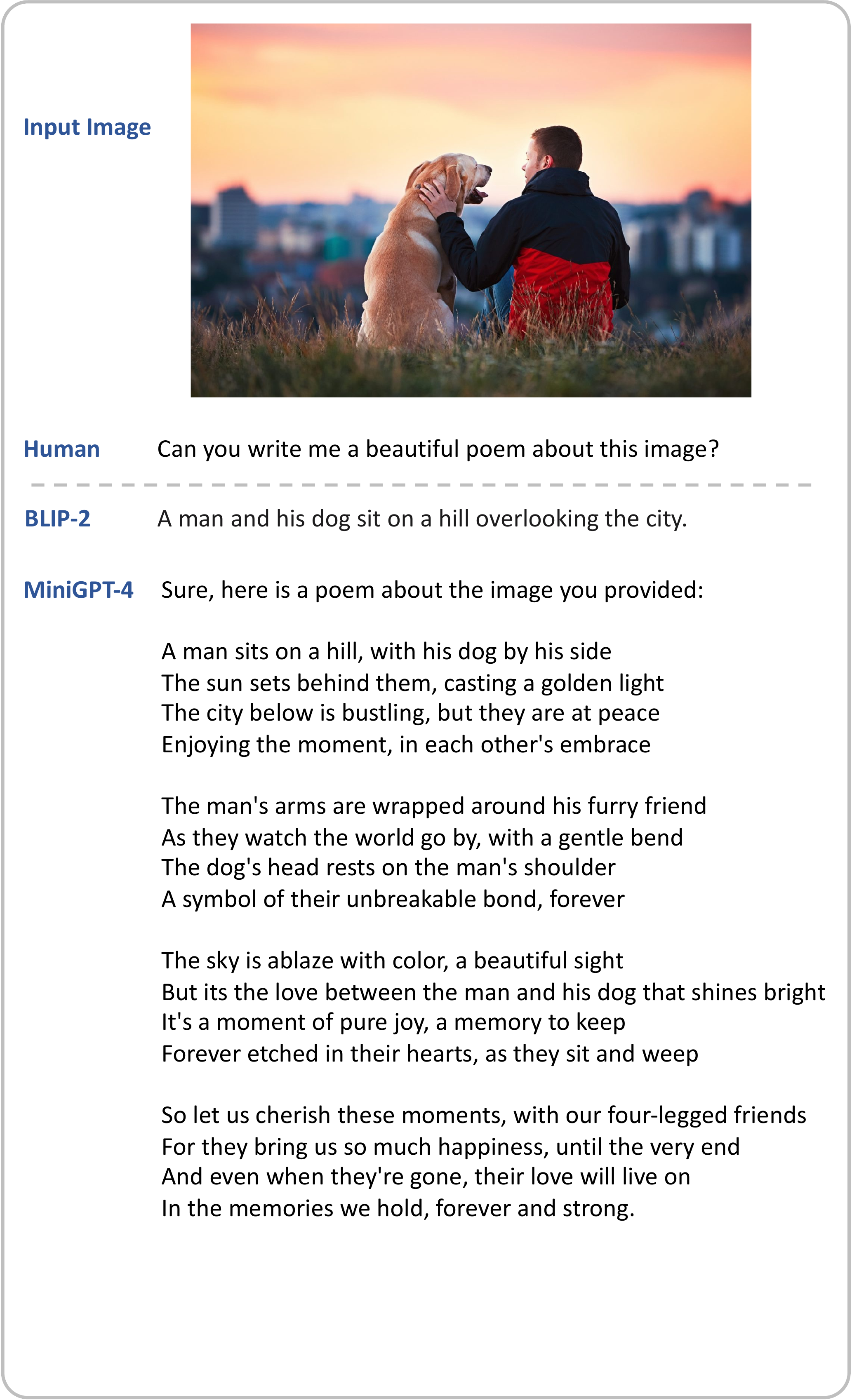}  
\caption{Poem writing}
\label{fig:poem}
\end{minipage} \hfill
\end{figure}

\begin{figure}[h]
\begin{minipage}{0.5\textwidth}
\centering
 \includegraphics[width=1.0\linewidth]
    {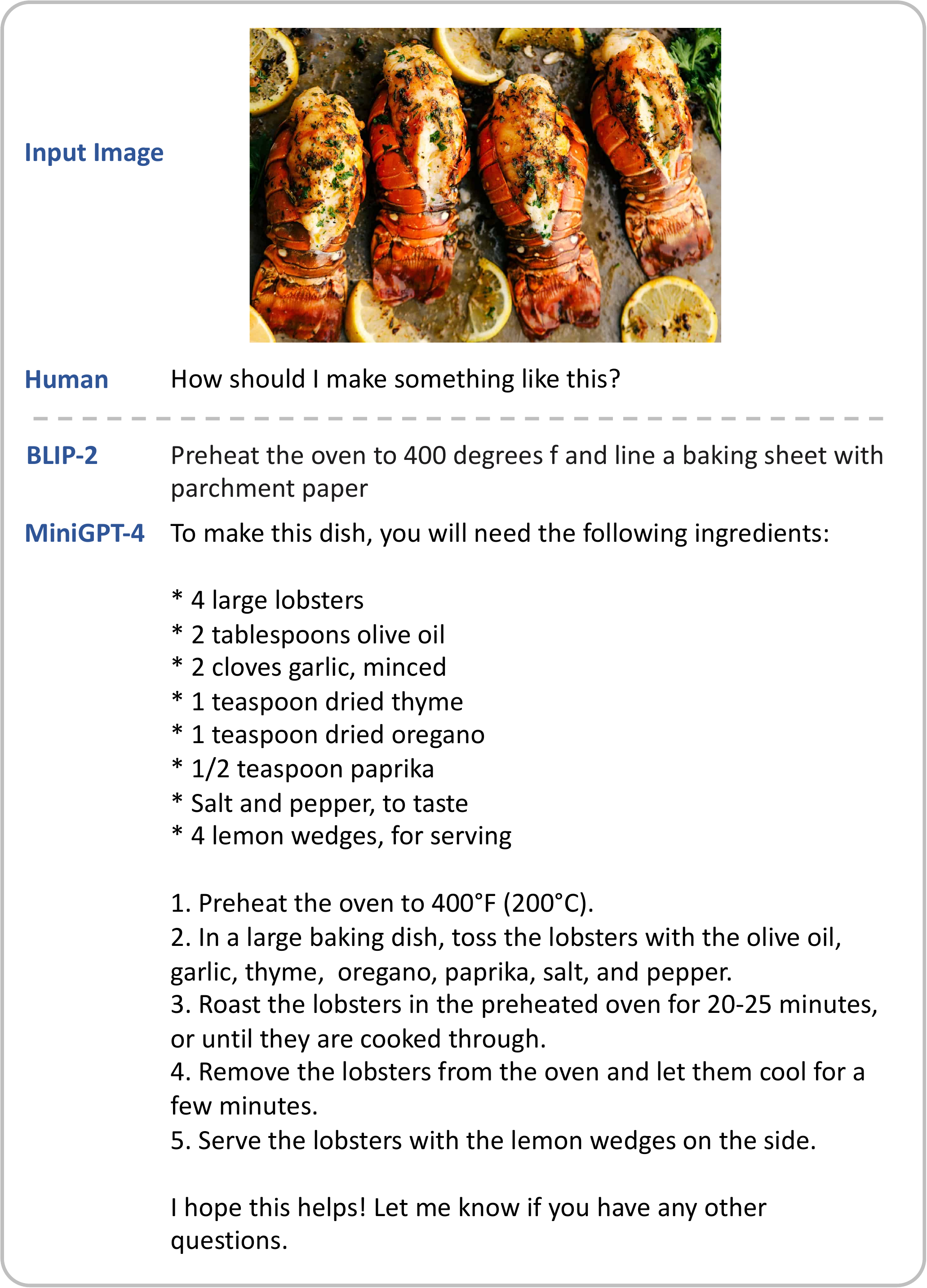}   
\caption{Food recipe generation}
\label{fig:cook}
\end{minipage} \hfill
\begin{minipage}{0.5\textwidth}
\centering
 \includegraphics[width=1.0\linewidth]
    {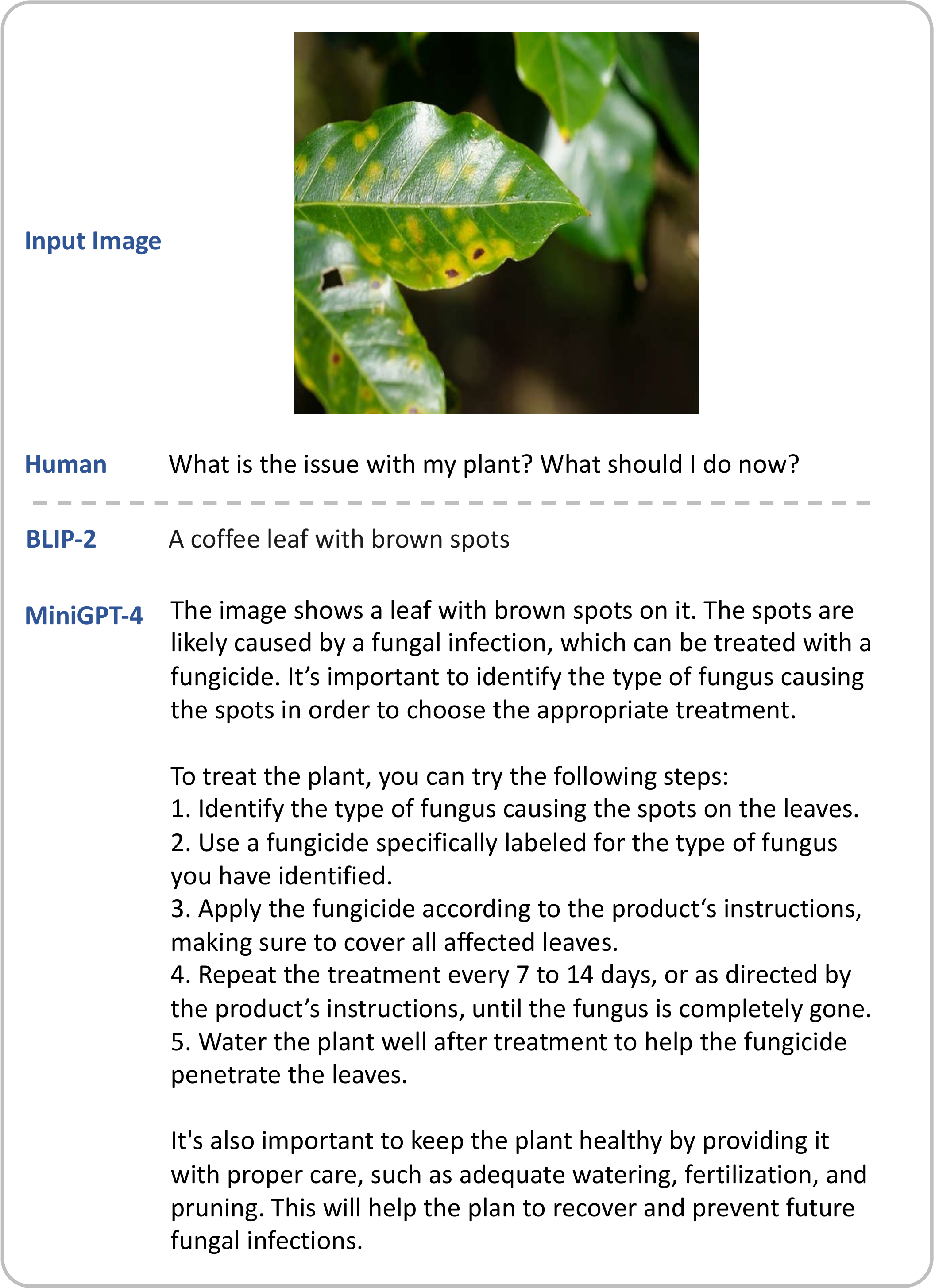}  
\caption{Plant cultivating}
\label{fig:plant}
\end{minipage} \hfill
\end{figure}

\subsection{Evaluation in traditional VQA benchmarks}
\label{appx: vqa}

The aim of this study is to replicate the remarkable multi-modal capabilities demonstrated in GPT-4, such as generating detailed image descriptions and creating websites from hand-drawn drafts.
To emphasize the most crucial component of advanced vision-language skills, the methodology of MiniGPT-4 is intentionally kept minimal.
For instance, the learnable model capacity is limited (only one linear layer), and MiniGPT-4 is trained with just 5 million pairs, in contrast to BLIP-2 with 129 million image-text pairs.
Such a pared-down approach is anticipated to yield suboptimal results on traditional benchmarks.
While this isn't our primary goal, we offer a quantitative analysis of the VQA datasets A-OKVQA (multi-choice) \citep{schwenk2022okvqa} and GQA \citep{hudson2019gqa}.
Additionally, to showcase the potential of MiniGPT-4 with traditional benchmarks, we conduct a straightforward ablation study. Here, we simply unfreeze the LLM using LoRA \citep{hu2021lora} and incorporate more training data from the VQAv2, OKVQA, and A-OKVQA datasets during the second finetuning stage.
Results in Tab.~\ref{tab_supp} indicate that the original MiniGPT-4 lags behind BLIP-2 by a reasonable margin, and merely augmenting the learning capacity and the training data results in a substantial performance improvement, which confirms our expectations.
We believe our model's performance on conventional vision benchmarks can be enhanced with a carefully designed training strategy (e.g., dataset sample ratios, learning rate schedule, etc.), more training data/datasets, and additional learnable parameters.
Since enhancing performance on traditional vision benchmarks isn't this project's objective, we reserve this aspect for future research.

\begin{table}[h!]
\centering
\scalebox{0.7}{
\begin{tabular}{l c c c}
\toprule
Model & Training data & AOK-VQA & GQA \\
\midrule
Blip-2 & 129M image-text pairs & 80.2 & 42.4 \\
MiniGPT-4 & 5M image-text pairs &  58.2 & 32.2 \\
MiniGPT-4 (Finetune Vicuna) & 5M image-text pairs & 67.2 & 43.5 \\
\bottomrule
\end{tabular}
}
\caption{Performance Comparison between BLIP-2 and MiniGPT-4}
\label{tab_supp}
\end{table}

\subsection{Details of Caption Evaluation}
\label{appx: caption_eval}
We employ ChatGPT to determine whether the baseline models cover all the objects and visual relations presented in the ground-truth captions. For the COCO evaluation dataset, we randomly choose one ground-truth caption and treat it as the reference caption. We apply the following prompt to perform the evaluation.

\textit{There is one image caption1 `\{ground-truth caption\}', and there is another image caption2 `\{comparison caption\}'. Does image caption2 cover all the objects and visual relations shown in image caption1? Only answer yes or no without any explanation.}

\subsection{More qualitative ablation results}

\begin{figure}[h]
\begin{minipage}{0.45\textwidth}
\centering
 \includegraphics[width=1.0\linewidth]
    {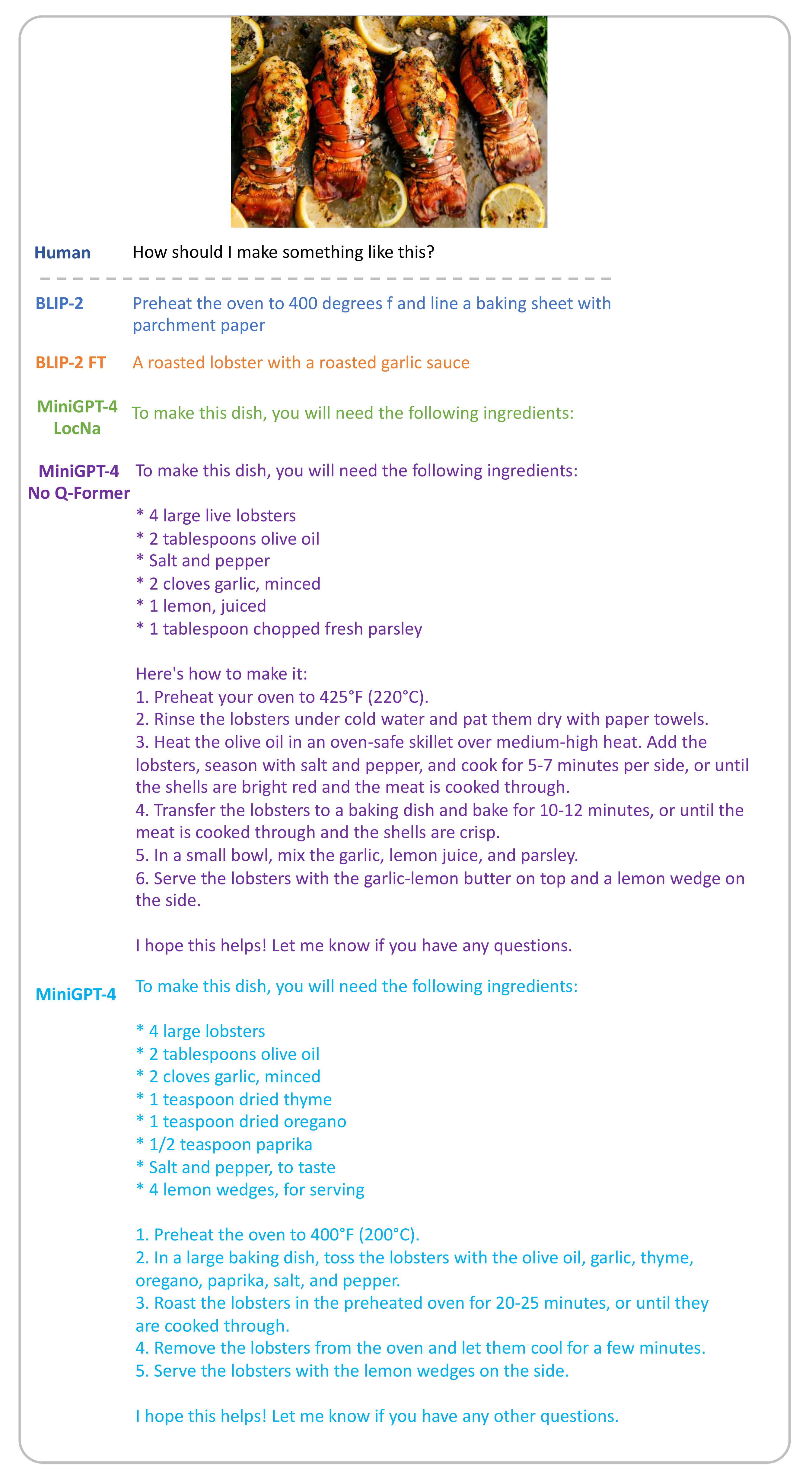}  
\caption{Ablation Study on Recipe Generation}
\label{fig:ab_cook}
\end{minipage} \hfill
\begin{minipage}{0.55\textwidth}
\centering
 \includegraphics[width=0.9\linewidth]
    {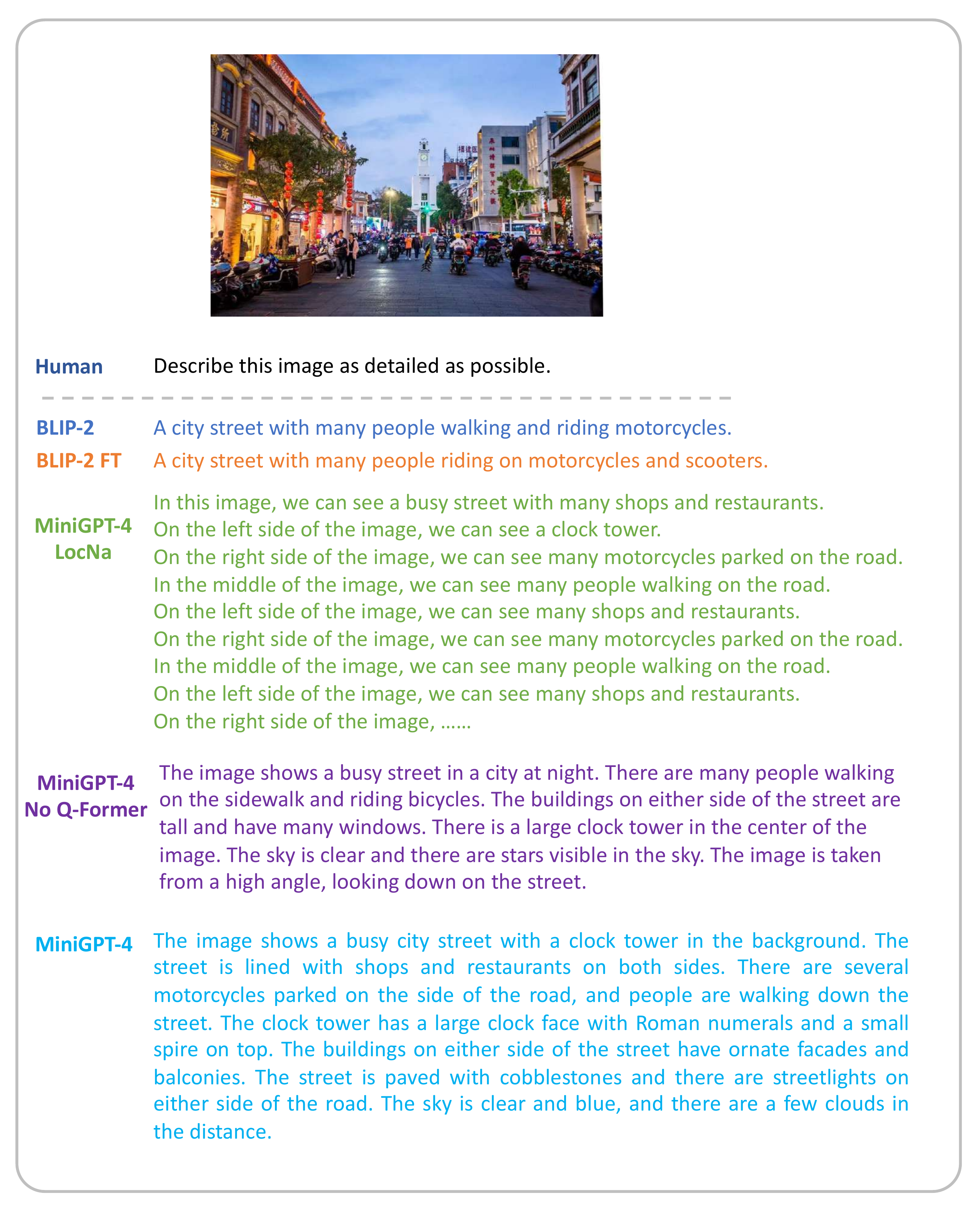}  
\caption{Ablation Study on Detailed Description}
\label{fig:ab_des}
\end{minipage} \hfill
\end{figure}

\end{document}